\documentclass{article}
\usepackage{ijcai19}
\usepackage{times}
\usepackage{soul}
\usepackage{url}
\usepackage[small]{caption}
\usepackage{booktabs}
\usepackage{mathtools}
\usepackage{color}
\usepackage{array}
\usepackage{stfloats}
\usepackage{amsthm}
\usepackage{float}
\usepackage{nccmath}
\usepackage{graphicx}
\usepackage[linesnumbered,ruled]{algorithm2e}
\usepackage{setspace}
\usepackage{booktabs}
\usepackage{mwe}
\usepackage{amsmath,amssymb,amsfonts}
\usepackage{graphicx}
\usepackage{textcomp}
\usepackage{xcolor}
\usepackage{xspace}
\usepackage{subcaption}
\newcommand{\BfPara}[1]{{\noindent\bf#1.}\xspace}
\newcommand{\argmin}{\mathop{\mathrm{argmin}}\limits} 
\newcommand{\argmax}{\mathop{\mathrm{argmax}}\limits} 

\title{Randomized Adversarial Imitation Learning for Autonomous Driving}

\author{MyungJae Shin$^1$\and Joongheon Kim$^1$\\ \affiliations $^1$Chung-Ang University, Seoul, Republic of Korea\\ \emails mjshin.cau@gmail.com, joongheon@gmail.com}

\begin{document}

\maketitle
\begin{abstract}
With the evolution of various advanced driver assistance system (ADAS) platforms, the design of autonomous driving system is becoming more complex and safety-critical. The autonomous driving system simultaneously activates multiple ADAS functions; and thus it is essential to coordinate various ADAS functions. This paper proposes a randomized adversarial imitation learning (RAIL) method that imitates the coordination of autonomous vehicle equipped with advanced sensors. The RAIL policies are trained through derivative-free optimization for the decision maker that coordinates the proper ADAS functions, e.g., smart cruise control and lane keeping system. Especially, the proposed method is also able to deal with the LIDAR data and makes decisions in complex multi-lane highways and multi-agent environments.

\end{abstract}

\section{Introduction}
With the increasingly growing interests in autonomous driving, the various forms of advanced driver assistance system (ADAS) functions such as smart cruise control (SCC), lane keeping system (LKS) and collision-avoidance systems (CAS) have been developed with high potentials in the enhanced convenience of drivers for limited on-driving situations. Especially, in multi-lane highway environments, it is essential to form efficient long term assistance strategies while maintaining safety because the malfunctions in safety cause on-road accidents and road congestion. The various ADAS functions presented in modern autonomous driving have high interdependence; thus it has to be regarded as a single integrated system. Therefore the strategies that properly coordinate the ADAS functions are required.

\begin{figure}[t!]
    \centering
    \includegraphics[width=0.9\columnwidth]{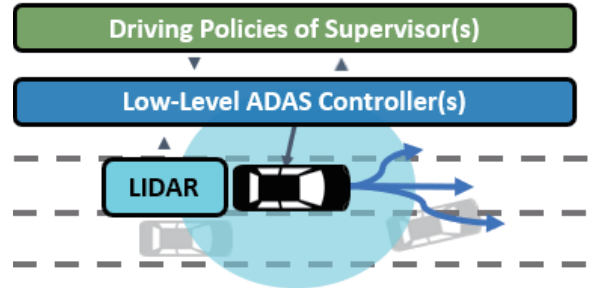}
    \caption{Simplified learning hierarchy to control vehicle systems.}
    \label{fig:overview}
    \vspace{-4mm}
\end{figure}

A conventional system hierarchy of autonomous vehicle is as illustrated in Fig.~\ref{fig:overview}. The low-level ADAS controllers are directly connected to the LIDAR sensors accessible in autonomous vehicle. The controllers determine the information needed to control the autonomous vehicle and transmit the determined operations to mechanical components. As a single integrated system, it is expected that multiple ADAS functions simultaneously cooperate to manage the systems operation of the vehicle. Therefore, a supervisor that coordinates the low-level controllers needs to select appropriate ADAS functions when the vehicle acts in dynamic on-road environments~\cite{korssen2018systematic}. The objective of the supervisor is to be a decision maker of the overall system during driving operation.
The problem is that the driving policies of the supervisor should satisfy the robustness regardless of various traffic environments. Prior research results on autonomous driving consist of diverse approaches with rule-based driving policies. However, these policies have been difficult to cope with time-varying environments (i.e., huge observation spaces and action spaces)~\cite{ahmed1999modeling}. Recently, the emergence of deep reinforcement learning (DRL), which utilizes powerful function approximations such as neural networks, allows the supervisor to obtain robust driving policies; it has made revolutionary progresses in the autonomous driving~\cite{mnih2015human,silver2016mastering,hoel2018automated,mukadam2017tactical}. However, there are challenges with DRL when the driving policies try to learn the policies that maximize the expected rewards during operation. The criteria for what should be the reward function of autonomous driving is still in progress in many studies. Furthermore, since there are undesirable policies to maximize the expected rewards at the expense of violating the implicit rules of the environments, it is difficult to learn the robust and safe policies through DRL in autonomous driving~\cite{pan2018agile}.
These problems motivate the researchers to adopt imitation learning (IL) to optimize the driving policy instead. The IL trains the driving policies based on the desired behavior demonstrations rather than the configuration of the specific reward functions as well as the IL can leverage domain knowledge. Based on the advantages of IL, it has been proved that the IL performs remarkably in the areas of robotics, navigation, autonomous vehicle, and etc~\cite{pomerleau1991rapidly,pomerleau1989alvinn,pan2018agile}.
However, the main challenge faced by many researchers is the techniques that combine DRL and IL algorithms require too much data to achieve reasonable performance, and the corresponding famous example is generative adversarial imitation learning (GAIL)~\cite{schulman2015trust,schulman2017proximal}. To address this issue, the algorithm models become complicated; and thus the models lead to reproducibility crisis. Furthermore, the models are sensitive to the implementation structure of the same algorithms and rewards from environments. For example, in GAIL, the discriminator of Generative Adversarial Networks (GAN) takes a role of the reward function. With the combination of discrimination and the complex DRL algorithms, e.g., TRPO and PPO, the GAIL trains the policies. As a result, the reconstruction results do not have always reasonable performance, and can stuck in sub-optimal even with marginal differences. These problems make the difficulties in training robust autonomous driving policies; the trained policies have not yet been successfully deployed to autonomous vehicles~\cite{henderson2017deep,islam2017reproducibility}. Recently, augmented random search (ARS) that consists of the natural gradient policy algorithm is proposed~\cite{rajeswaran2017towards}. Because the ARS is a derivative-free simple linear policy optimization method, it is relatively easy to reconfigure the robust trained policy that shows reasonable performance.
In this work, we present an IL-based method that combines the concepts of ARS and GAIL. 
For more details, random search based \textit{randomized adversarial imitation learning (RAIL)} algorithm is proposed in this paper; and the RAIL algorithm trains policies using randomly generated matrices where the random matrices are used to search update directions that lead to optimal policies.
This approach is advantageous in terms of computation (such as back-propagation) overhead reduction whereas DRL algorithms that use gradients to optimize weights.
Furthermore, by leveraging expert demonstrations, our system can learn the driving policies of supervisor that achieves similar performance compared to the expert in terms of average speeds and lane changes. Through the data-intensive performance evaluation, it is demonstrated that the proposed RAIL algorithm can train the autonomous driving decision maker as desired.

\BfPara{Contributions}
Our proposed RAIL method shows that the random search in the space of policy parameters can be adapted to IL for autonomous driving policies. For more details, our contributions are as follows:
(i) self-driving mechanism is proposed inspired by IL. Our method can successfully imitate expert demonstrations; and the corresponding static and linear policies can achieve similar speeds with many lane changes and overtakes.
(ii) previous IL methods are based on conventional RL methods which show complicate configurations to control autonomous driving. However, RAIL has simplicity based on derivative-free random search.
(iii) this method has not been previously applied to learn the robust driving policies in autonomous driving.

\BfPara{Organization}
Sec.~\ref{sec:related} and Sec.~\ref{sec:background} describe the previous work and background knowledge. 
Sec.~\ref{sec:pe} defines our problem, i.e., training policies for autonomous driving. Sec.~\ref{sec:rail} designs the RAIL algorithm. Sec.~\ref{sec:experiment} shows the experiment results based on expert demonstrations in highway autonomous vehicle control environments. Sec.~\ref{sec:conclusion} concludes this paper.

\section{Related Work}\label{sec:related}
\BfPara{Imitation learning (IL)} The IL methods are divided into two categories, i.e., behavioral cloning (BC) and inverse reinforcement learning (IRL). The BC is considered as the simplest IL method. To restore expert policy, it works by collecting training data from the expert driver’s behaviors, and then uses it to directly learn the corresponding policy. If the policy deviates from trajectories that is trained in the training procedure, the agent tends to be fragile. This is because behavior cloning tries to reduce the 1-step deviation error of training data, not to reduce the error of entire trajectories. Prerequisites for reasonable policy restoration is a sufficient number of expert driving demonstrations. On the other hand, IRL has an intermediate procedures to estimate and recover the hidden reward function which explains the expert demonstration~\cite{ziebart2008maximum,finn2016guided}. Since IRL has to optimize the policy as well as the reward function, IRL generally implies significant computational costs. In \cite{finn2016connection} an  d \cite{ho2016generative}, the theoretical and practical considerations of connections between IRL and adversarial network is studied. GAIL framework learns a policy that can imitate expert demonstration using the discriminator network, which bypasses the reward function optimization. 

\BfPara{Simplest model-free RL} The simplest model-free RL method that can solve standard benchmarks of RL has been studied under the two different directions: linear policies via natural policy gradients~\cite{rajeswaran2017towards} and a derivative-free policy optimization~\cite{es}. \cite{rajeswaran2017towards} shows that complicated structures of policies are not needed to solve continuous control problems. The authors train linear policies via natural policy gradients. The trained policies obtain competitive performance on the complex continuous environments. In\cite{es}, the authors showed that evolution strategies (ES) offers less data efficiency than traditional RL, but offers many advantages. Especially, a derivative-free optimization allows ES to be more efficiently in distributed learning. Furthermore, the trained policies tend to be more diverse compared to policies trained by traditional RL methods. In \cite{ars}, the connection between \cite{es} and \cite{rajeswaran2017towards} is studied to obtain the simplest model-free RL method yet, a derivative-free optimization for training linear policies. The proposed simple random search method showed state-of-art sample efficiency compared to competing methods in MuJoCo locomotion benchmarks.

\section{Background}\label{sec:background}
\subsection{Markov Decision Process (MDP)}
MDP is formalized by $(S, A, p(s), p(s'|s,a), r(s,a,s'), \gamma)$ where $S$, $A$, $p(s)$, $p(s'|s,a)$, $r(s, a, s')$, and $\gamma$ stand for set of states, set of actions, initial state distribution, environmental dynamic represented as conditional state distribution, reward functions, and discount factor, respectively. 
The environment interactions between a subject and its environment is unbounded in the continuing tasks; and thus the returns are defined as $R_t = \sum_{i=t}^{\infty}{\gamma^{i-t}r(s_i,a_i,s_{i+1})}$. The objective of MDP is to find a policy that maximizes the expected returns.

\subsection{Generative Adversarial IL (GAIL)}
GAIL is used for reward function in this paper. Based on GAN, the GAIL trains a binary classifier, $D(s,a)$, referred to as the discriminator, to distinguish between transitions sampled from an expert demonstration and those generated by the trained policies. With GAIL, an agent is able to learn a policy that imitates expert demonstrations using the adversarial network. The objective of GAIL is defined as follows: 
  \begin{multline}
  \label{eq:lossGAIL}
     \argmin_\theta \argmax_\phi \left\{\mathbb{E}_{\pi_\theta} \left[ \log \mathcal{D}_\phi(s, a)\right] +\right. \\ \left. \mathbb{E}_{\pi_E} \left[ \log(1-\mathcal{D}_\phi(s,a))\right]
     - \lambda H(\pi_\theta)\right\}.
  \end{multline}
where $\pi_\theta, \pi_E$ are the policy which is parameterized by $\theta$ and an expert policy.
In \eqref{eq:lossGAIL}, $H(\pi_\theta)\triangleq\mathbb{E}_{\pi}\left[-\log\pi(a|s)\right]$ is entropy regularization. $\mathcal{D}_\phi(s, a) \to \left[ 0, 1 \right]$ is the discriminator parameterized by $\phi$~\cite{ho2016generative}. In GAIL, the policy is instead provided a reward for confusing the discriminator, which is then maximized via some on-policy RL optimization schemes. The $\mathcal{D}_\phi$ takes the role of a reward function; and thus it gives learning signal to the policy~\cite{ho2016generative,gasil,optiongan}. 

\subsection{Augmented Random Search (ARS)}
ARS is a model-free RL algorithm. Based on random search in the parameter spaces of policies, ARS uses the method of finite differences to adjust its weights and learn the way how the policy performs its given tasks~\cite{matyas1965random,ars}. Through the random search in the parameter spaces, the algorithm can conduct a derivative-free optimization with noises~\cite{matyas1965random,ars}. To update the weights effectively, ARS selects update directions uniformly and updates the policies along with the selected direction. For updating the parameterized policy $\pi_\theta$, the update direction is as $\frac{r(\pi_{\theta - \nu\delta}) - r(\pi_{\theta + \nu\delta})}{\nu}$ where $\delta$ is a zero mean Gaussian vector, $\nu$ is a positive real number which represents the standard deviation of exploration noise, and $r(\pi_\theta \pm \nu \delta)$ means the rewards from environments when the parameter of policies is $\pi_\theta \pm \nu\delta$.
Let $\theta_t$ be the weight of policy at $t$-th training iteration. $N$ denotes that the number of sampled directions per iteration. The update step is configured as:
\begin{equation}
    \label{brs}
    \theta_{t+1} = \theta_{t} + \frac{\alpha}{N} \sum_{i=1}^{N}\nolimits{\left[r(\pi_{\theta + \nu\delta_i}) - r(\pi_{\theta - \nu\delta_i})\right]\delta_i}.
\end{equation}

However, the problem of random search in the parameter spaces of policies is large variations in terms of the rewards $r(\pi_\theta \pm \nu\delta)$ which are observed during training procedure. The variations make the updated policies to be perturbed through the update steps~\cite{ars}. To address the large variation issue, the standard deviation $\sigma_R$ of the rewards which is collected at each iteration is used to adjust the size of the update steps in ARS. Based on the adaptive step size, ARS shows better performance compared to DRL algorithms (i.e., PPO, TRPO, etc.) in specific environments.

\section{Problem Definition}\label{sec:pe}
\BfPara{Motivation}
By coordinating the ADAS functions in the limited situations such as highways, the autonomous driving can be realized. To coordinate the ADAS functions for autonomous driving, the supervisor determines the appropriate ADAS functions based on the nearby situations. However, the complete states of the environment are not known to the autonomous vehicle supervisor. The supervisor receives an observation that is conditioned on the current state of the system. The host vehicle interacts with the environment including surrounding vehicles and lanes; and thus it uses partially observable local information. Therefore, we need to model the observation of agent as an $(O, A, T, R, \gamma)$ tuple representing a partially observable Markov decision process with continuous observations and actions for autonomous driving. Similar to MDP in Section \ref{sec:background}, there are the set of partial observation states denoted by $O$, instead of $S$. In this paper, LIDAR data is regarded as the observation by vehicles. 

In this paper, a finite state space $\mathcal{O} \in \mathbb{R}^n$ and a finite action space $\mathcal{A} \in \mathbb{R}^p$ are considered. The goal of IL for autonomous driving is to learn a policy $\pi_\theta  \in \Pi : \mathcal{O} \times \mathcal{A} \to \mathbb{R}^p$ which imitates expert demonstration from GAN $\mathcal{D}_\phi(s, a) \to \left[ 0, 1 \right]$ where $\theta \in \mathbb{R}^n$ are the policy parameters and $\phi \in \mathbb{R}^{n+p}$ are the discriminator parameters~\cite{ho2016generative}.

\BfPara{The state space} For sensor model, we use a vector observation that consists of LIDAR sensor data. In particular, $N$ beams are spread evenly over the field of view $\left[ \omega_{min}, \omega_{max} \right]$. The LIDAR sensor detects around the vehicle. Each sensor data has a maximum range of $r_{max}$. The sensor returns the distance between the first obstacle it encounters and the host vehicle, or $r_{max}$ if no obstacle is detected. Then, the observation is described as $O = (o_1,\dots,o_N )$. Furthermore, based on the distance information, the relative speed of the obstacle and the host vehicle can be calculated. Here, the relative speed observation is described as $V_r = (v_1,\dots,v_N)$.

\BfPara{The action space} 
The policy is considered as a high-level decision maker which determines optimal actions based on observation on the highway. We assume that the autonomous vehicle utilizes the ADAS functions; and thus the determined actions of driving policy activate each ADAS function. The driving policy is defined in a discrete action space. The high level decisions can be break down into the following 5 actions as follows: (1) maintain current status, (2) accelerate speed for a constant amount $vel_{cur} + vel_{acc}$, (3) decelerate speed for a constant amount $vel_{cur} - vel_{dec}$, (4) make a left lane change, (5) make a right lane change. The actions expect that the vehicle is adjusted with autonomous emergency braking (AEB) and adaptive cruise control (ACC)~\cite{mukadam2017tactical,min2018deep,hoel2018automated}. 

\BfPara{The reward function}
In the GAIL framework, the reward from adversarial network is defined $r_{\pi_\theta}(s, a) = - \log(1 - \mathcal{D}_\phi(s, a))$ or $r_{\pi_\theta}(s, a) = \log(\mathcal{D}_\phi(s,a))$~\cite{ho2016generative}. 
The former type of the reward is used to encourage agent to train survival policies through a survival bonus in the form of positive reward based on their lifetime. The latter is often used to train policies with a per step negative reward, when a reward function consists of a negative constant for the state and action. However, in this case, it is hard to learn the survival policies~\cite{kostrikov2018discriminator}. The prior knowledge of environmental objectives is important, but the environment-dependent reward function is undesirable when the agent requires interactions with a training environment in order to imitate an expert policy. Therefore, we defined the reward function as follows: $\log(\mathcal{D}_\phi(s,a))- \log(1 - \mathcal{D}_\phi(s, a))$.

\section{Randomized Adversarial IL (RAIL)}\label{sec:rail}
\vspace{-0.1mm}

\begin{figure*}[t!]
    \centering
    \includegraphics[width=2.0\columnwidth]{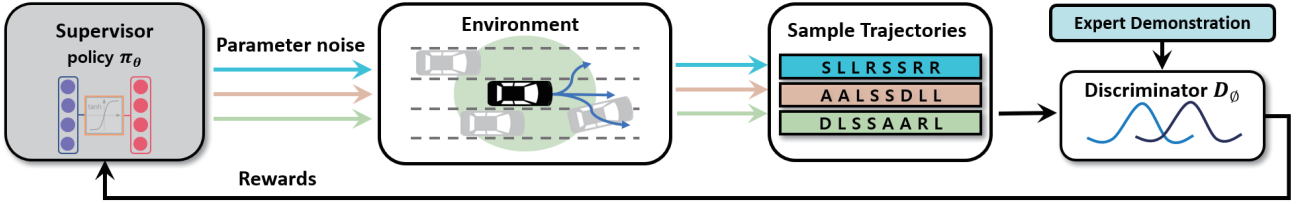}
    \caption{Structure of RAIL.}
    \label{fig:structure}
\end{figure*}

The approach in this paper, named \textit{randomized adversarial imitation learning (RAIL)}, adopts IL through adversarial network paradigm (i.e., GAIL). The main concept of RAIL is enhance an conventional algorithm called ARS and GAIL~\cite{ho2016generative,ars}. RAIL aims to train the driving policy $\pi_\theta$ to imitate expert driver's demonstration. This section describes the details of RAIL and makes a connection between GAIL and a derivative-free optimization.  

In Fig. \ref{fig:structure}, the overall structure of RAIL is described. The supervisor of host vehicle is considered as an agent which has policy $\pi_\theta$. From the environment (i.e., multi-lane highway), the host vehicle receives the observation. Then, the random noise matrices of small values can be generated. The noise matrices are added to or subtracted from the policy parameters $\theta$. As a result, several different temporary policies are produced. The agent interacts with the environments multiple times based on the generated noisy policies and the results are collected as sample trajectories. Based on the samples, the main policy $\pi_\theta$ is trained to control the autonomous driving successfully with fully utilizing ADAS functions which guarantee safety. In the training process, the policy $\pi_\theta$ attempts to fool a discriminator $\mathcal{D}_\phi$ into believing the sample trajectories of the agent come from expert demonstrations. The $\mathcal{D}_\phi$ tries to distinguish between the distribution of trajectories which are sampled by the policies $\pi_\theta$ and the expert trajectories $\mathcal{T}_E$. The trajectories consist of state-action pair $(s, a)$. The discriminator takes the role of the reward module in RAIL, as shown in Fig. \ref{fig:structure}; and thus the policy $\pi_\theta$ is trained against the discriminator. Therefore, the performance of the discriminator has a significant impact on convergence and agent.

As shown in Fig.\ref{fig:structure}, the discriminator is trained based on sample trajectories and expert demonstration. However, in training procedure, since the policy $\pi_\theta$ is updated every iteration, the distribution of the sample trajectories changes. As a result, the training of the discriminator is not stabilized; and thus it gives the inaccurate reward signal to the policy. Consequently, the policy can be perturbated during update step~\cite{gasil}. 
In RAIL, the loss function of least square GAN (LS-GAN) is used to train a discriminator $\mathcal{D}_\phi$~\cite{mao2017least}, and the objective function of the discriminator is as follows:
\begin{equation}
  \label{lsgan_loss}
  \begin{multlined}
     \argmin_\phi L_{LS}(\mathcal{D}) = \frac{1}{2}\mathbb{E}_{\pi_E} \left[(\mathcal{D}_\phi(s, a) - b)^2\right]+\\
     \frac{1}{2}\mathbb{E}_{\pi_\theta} \left[ (\mathcal{D}_\phi(s,a) - a)^2\right]
\end{multlined}
\end{equation}
where $a$ and $b$ are the discriminator labels for the sampled trajectories from the policy $\pi_\theta$ and the expert trajectories. 

In this paper, least-squares loss function is used to train the discriminator. When the loss function of original GAN Eq.\ref{eq:lossGAIL} is used, sampled trajectories which are far from the expert trajectories but on the correct side of the decision boundary are almost not penalized by sigmoid cross-entropy loss. In a contrast, the least-squares loss function (\ref{lsgan_loss}) penalizes the sampled trajectories which are far from the expert trajectories on either side of decision boundary~\cite{mao2017least}. Therefore, the stability of training is improved; and it leads the discriminator to give accurate reward signals to the update step. In LS-GAN, $a$ and $b$ have relationship $b-a=2$ for \eqref{lsgan_loss} to be Pearson $\mathcal{X}^2$ divergence~\cite{mao2017least}. However, we use $a=0$ and $b=1$ as the target discriminator labels. The results of the discriminator $\mathcal{D}_\phi$ are in the range of 0 to 1 (experimentally determined). In RAIL, the discriminator is interpreted as a reward function for policy optimization. Forementioned in Sec.\ref{sec:pe}, the form of reward signal is as follows:
\begin{equation}
  \label{reward_signal}
  \begin{multlined}
     r_{\pi_\theta}(s, a) = \log(\mathcal{D}_\phi(s,a))- \log(1 - \mathcal{D}_\phi(s, a))
\end{multlined}
\end{equation}

This means that if the trajectories sampled from the policy $\pi_\theta$ is similar to expert trajectories, the policy $\pi_\theta$ gets higher reward $r_{\pi_\theta}(s, a)$. The policy $\pi_\theta$  is updated to maximize the discounted sum of rewards given by the discriminator rather than the reward from the environment as shown in Fig.~\ref{fig:structure}. The objective of RAIL can be described as $\argmax_{\theta} \mathbb{E}_{(s, a)\thicksim \pi_\theta}\left[ r(s, a)\right]$, and then, it is as follows by \eqref{reward_signal}:
\begin{equation}
  \label{RAIL_objective}
    \argmax_{\theta}    \mathbb{E}_{(s, a)\thicksim \pi_\theta}\left[ \log(\mathcal{D}_\phi(s,a))- \log(1 - \mathcal{D}_\phi(s, a)) \right]
\end{equation}
where this \eqref{RAIL_objective} represents the connection between adversarial IL and randomized parameter space search in RAIL.

\begin{algorithm}[t]
\small
    \SetKwInOut{Hyperparameters}{Hyperparameters}
    \SetKwInOut{Initialize}{Initialize}
    \Hyperparameters{$\alpha$ step size, $N$  number of sampled directions per iteration, $\delta^i$ and $\delta^o$ Gaussian vectors from zero mean and $\nu$ a positive real number standard deviation of the exploration noise, $h$ hidden layer size}
    \Initialize{$\theta^i$, $\theta^o$ from behavior cloning, $\mu_0 = \bold{0}\in \mathbb{R}^n,$ and $\sum_{0} = \bold{I}_n \in \mathbb{R}^{n\times n}$}
    
    \While{$t \le$ Episode Length}
    {
        i.i.d. Random Sampling with $\bold{\delta_t} = \left\{ \delta_1, \delta_2, . . . , \delta_N ; \delta^i_k \in \mathbb{R}^{h \times n}, \delta^o_k \in \mathbb{R}^{p \times h} \right\} $ \\
        Collect $2N$ rollouts and corresponding rewards using the $2N$ noisy policies for $k\in \left\{ 1,2,\dots,N \right\}$.\\
        
        \Indp 
        $\pi_{t,(k),+} = (\theta_t + \nu\delta_k)$diag$(\sum_t)^{-1/2}(s - \mu_t )$\\
        $\pi_{t,(k),-} = (\theta_t - \nu\delta_k)$diag$(\sum_t)^{-1/2}(s - \mu_t )$\\
        \Indm 
        
        Update discriminator parameter $\phi_t$ :\\
        
        $\nabla_{\phi_t}L_{LS} = \frac{1}{2}\mathbb{E}_{\pi_E} \left[(\nabla_{\phi_t}\mathcal{D}_{\phi_{t}}(s, a) - b)^2\right]$\\ 
        \Indp\Indp\Indp\Indp
        $+\frac{1}{2}\mathbb{E}_{\pi_\theta}\left[(\nabla_{\phi}\mathcal{D}_{\phi_{t}}(s,a) - a)^2\right]$\\
        \Indm\Indm\Indm\Indm
         
        Update the policy parameter $\theta_t$ :\\
        
        $\theta_{t+1} = \theta_{t} + \frac{\alpha}{N\sigma_R}\sum^{N}_{i=1}{\left[ r(\pi_{t,(k),+}) - r(\pi_{t,(k),-}) \right]\delta_{(k)} }$\\

        where trajectories $T$ sampled from $\pi_{(t,(k),\pm)}$ \\ 
        $r(\pi_{t,(k),\pm}) = \mathbb{E}_{(s,a)\thicksim \pi_{t,(k),\pm}} \left[ \log(\mathcal{D}_\phi(s,a))- \log(1 - \mathcal{D}_\phi(s, a)) \right]$\\
        Set $\mu_{t+1}$, $\sum_{t+1}$ to be the mean and covariance of the states encountered from the start of training.\\
        $t$ = $t + 1$
    }
    \caption{RAIL}
    \label{algo:RAIL}
\end{algorithm}
\BfPara{Algorithm}
As mentioned, RAIL is related to ARS which is model-free reinforcement algorithm. Thus, RAIL utilizes parameter space exploration for a derivative-free policy optimization. The parameters of $\pi_\theta$ are denoted by $\theta$. The $\pi_\theta$ consists of $\pi^i_\theta$, $\pi^o_\theta$, and activation function. The $\pi^i_{\theta^i}$ is the input layer of $\pi_\theta$ where $\theta^i \in \mathbb{R}^{n\times h}$ are the parameters of the input layer. In addition, $\pi^o_{\theta^o}$ is the output layer where $\theta^o \in \mathbb{R}^{h\times p}$. The noises $\delta^i$ and $\delta^o$ of parameter space for exploration are $n \times h$ and $h \times p$ matrices where they are sampled from zero mean and $\nu$ standard deviation Gaussian distribution. In this paper, let $\theta$ be a set of $\theta^i$ and $\theta^o$. $\delta$ means a set of $\delta^i$ and $\delta^o$.

The pseudo-code of RAIL is represented in Algorithm~\ref{algo:RAIL}. The policy parameters $\theta^i$ and $\theta^o$ are initialized from behavior cloning. In training procedure, the noises $\bold{\delta^i} and \bold{\delta^o}$ which mean the search directions in parameter space of policy are chosen randomly for each iteration (line 2). Each set of selected $N$ noises makes two policies in the current policy $\pi_\theta$. We collect $2N$ rollouts and rewards from $N$ noisy policies $\pi_{t,k,\pm} = \theta_{t} \pm \nu\delta_k$ (line 3-6). The high dimensional complex problems have multiple state components with various ranges; and thus it makes the policies to result in large changes in the actions when the same sized changes are not equally influence state components. Therefore, the state normalization is used in RAIL (line 4-5,14); and it allows policy $\pi_{t,i,\pm}$ to have equal influence for the changes of state components when there are state components with various ranges~\cite{ars,es,nagabandi2018neural}. The discriminator $\mathcal{D}_\phi$ gives the reward signal to update step. However, since the trajectories for the training of the discriminator can only be obtained from current policies $\pi_{\theta_t}$, a discriminator is trained whenever the policy parameter $\theta_t$ is updated. The discriminator $\mathcal{D}_\phi$ finds the parameter $\phi$ which minimizes the objective function (\ref{lsgan_loss}) (line 7-9). By using the reward signals from the discriminator, the policy weight is updated in the direction of $+\delta$ or $-\delta$ based on the result of $r(\pi_{t,(k),+}) - r(\pi_{t,(k),-})$ (line 10-13). The state normalization is based on the information of the states encountered during the training procedure; and thus $\mu$ and $\sum$ are updated (line 14).

\section{Experiments}\label{sec:experiment}
In this section, we compare the performance between RAIL and baselines. Furthermore, in order to assess the performance gaps between the single-layer and multi-layer policies trained by RAIL, the single-layer and two-layer (i.e., multi-layer) policies was implemented. 

\BfPara{Simulator}
The simulated road environment is a highway driving roadway composed of five lanes. Other vehicles are generated in the center of the random lanes within a certain distance to the host vehicle. In addition, it is assumed that other vehicles do not collide with others while randomly changes the lanes. Aforementioned in Sec.~\ref{sec:pe}, the observation is based on LIDAR sensor result. We assume that LIDAR sensor detects a range of $360$ degrees with one ray per $15$ degree. The ray returns the distance between the first obstacle it encounters and the host vehicle. If there are no obstacles, it returns the maximum sensing range. We make the expert demonstration using PPO with specific action controls. The results present the average of 16 experimental results. In the experiments, the trained weights through BC are used to fast convergence in GAIL and RAIL. This simulation study is inspired by \cite{min2018deep}. We implemented the RAIL simulator based on Unity.

\begin{figure*}[t!]
   \centering
   \begin{subfigure}{.33\textwidth}
    \centering
    \includegraphics[width=1\columnwidth]{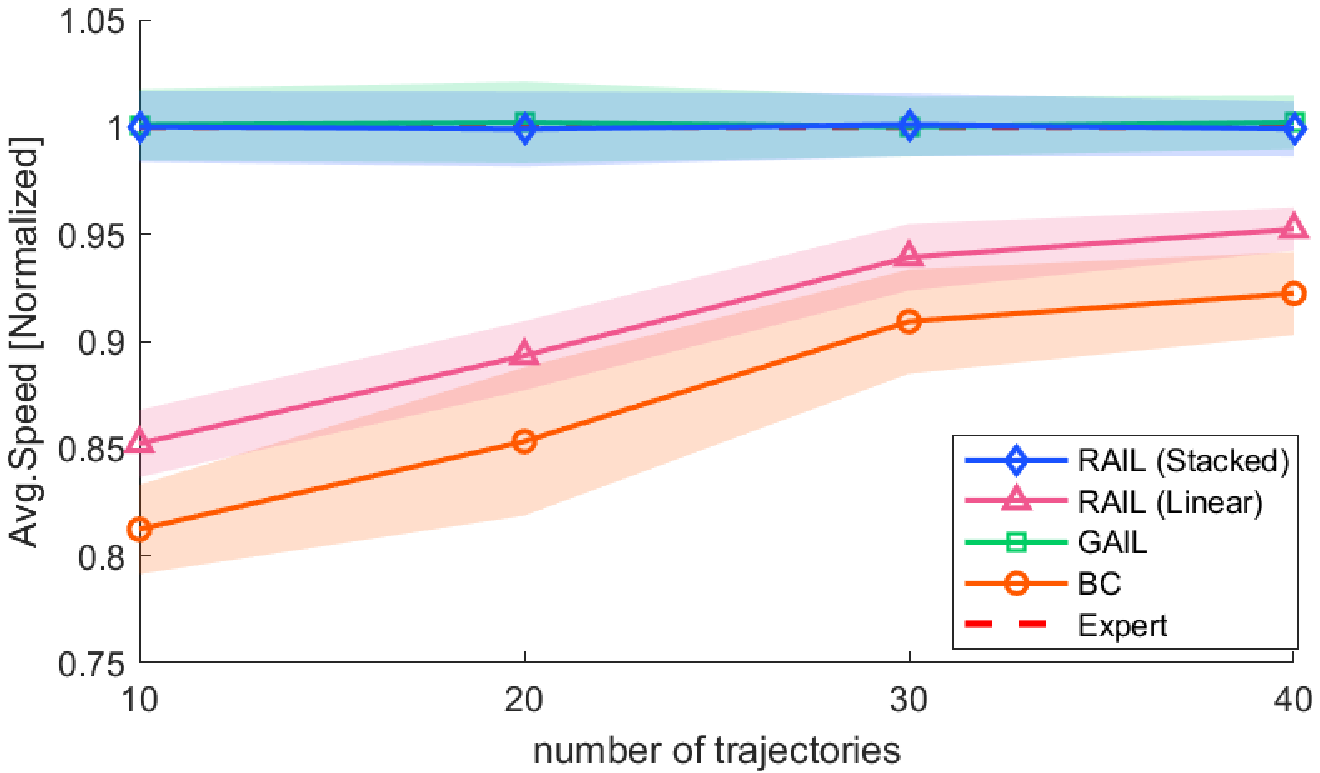}
    \caption{Normalized Speed.}
    \label{fig:speed}
  \end{subfigure}
   \begin{subfigure}{.33\textwidth}
    \centering
    \includegraphics[width=1\columnwidth]{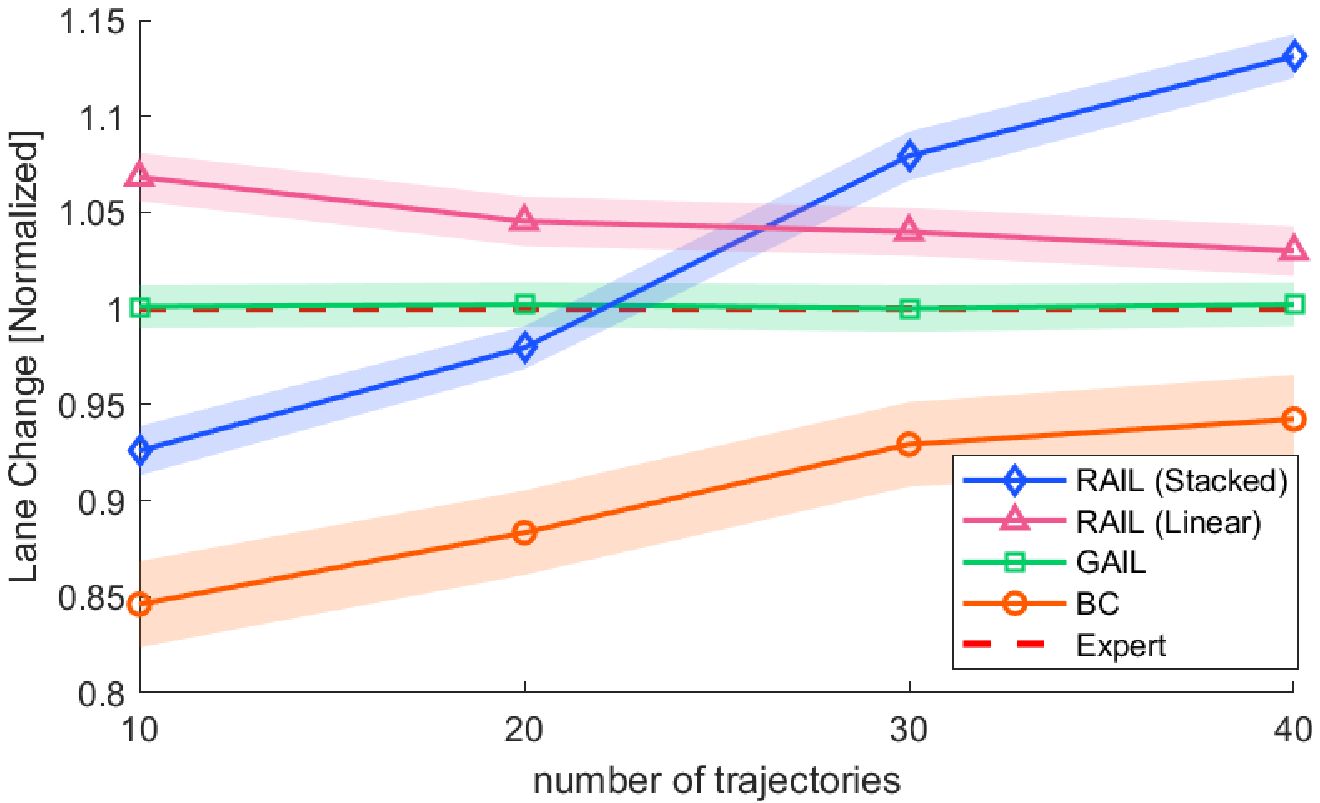}
    \caption{Normalized Lane change.}
    \label{fig:lane}
  \end{subfigure}
   \begin{subfigure}{.33\textwidth}
    \centering
    \includegraphics[width=1\columnwidth]{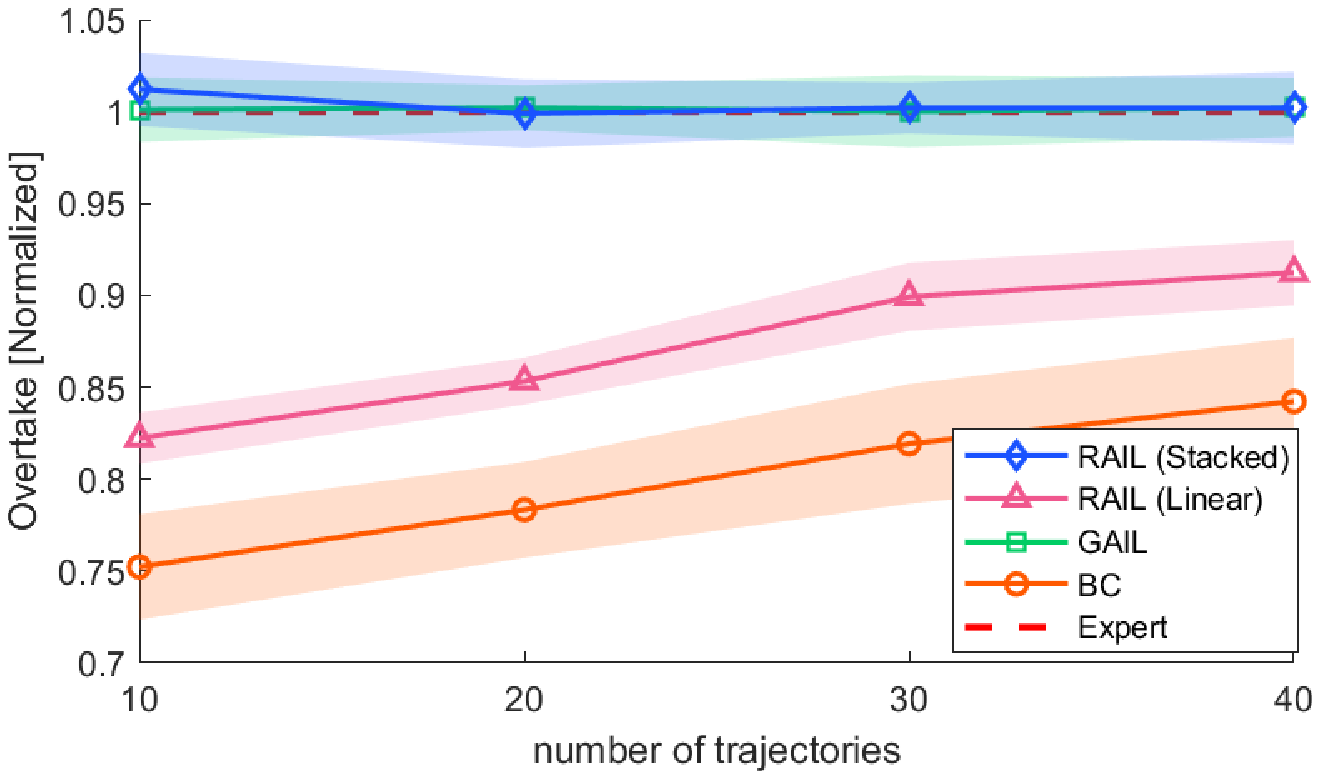}
    \caption{Normalized Overtake.}
    \label{fig:overtake}
  \end{subfigure}
  \caption{The performance of trained policy according to the set number of expert trajectories (Average of 5 episodes).}
  \label{fig:sample}
\end{figure*}

\begin{figure*}[t!]
   \centering
   
   \begin{subfigure}{.33\textwidth}
    \centering
    \includegraphics[width=1\columnwidth]{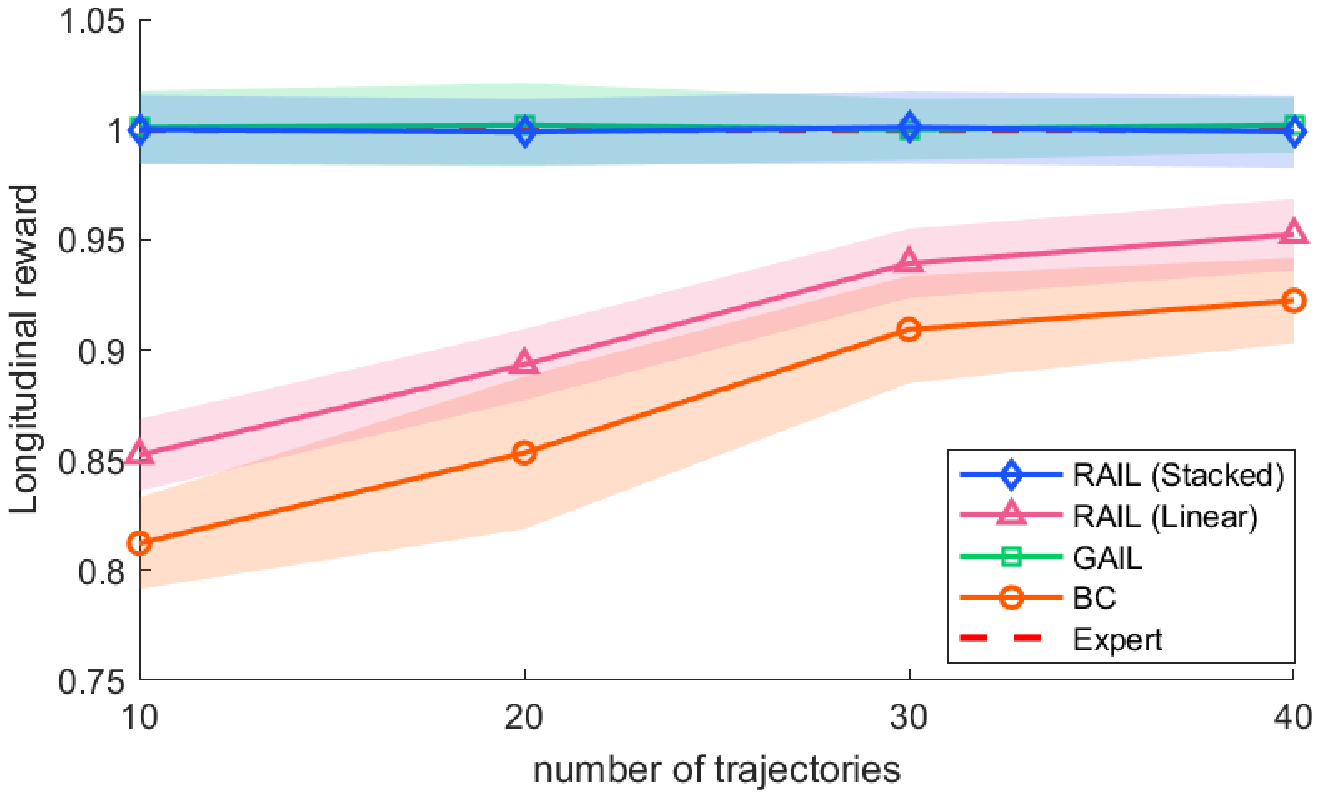}
    \caption{Normalized Longitudinal Rewards.}
    \label{fig:long}
  \end{subfigure}
   \begin{subfigure}{.33\textwidth}
    \centering
    \includegraphics[width=1\columnwidth]{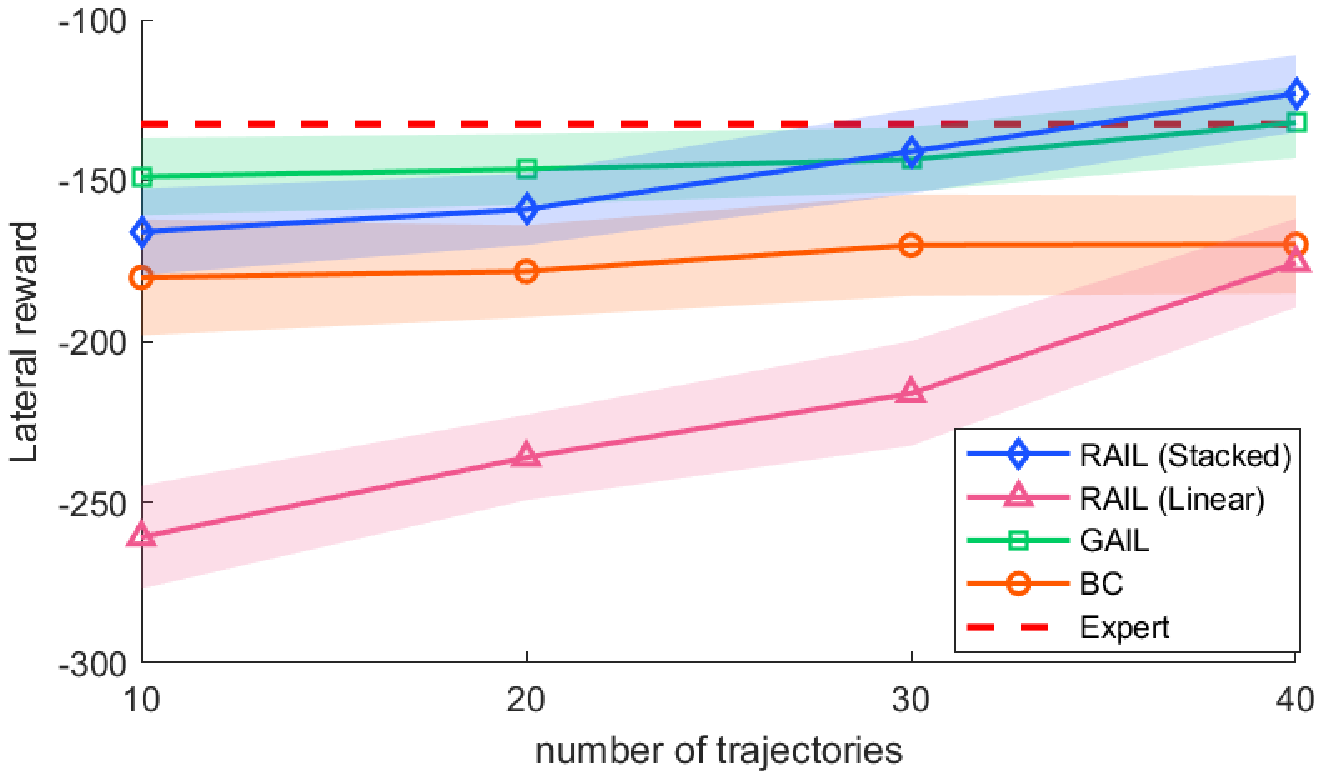}
    \caption{Lateral Rewards.}
    \label{fig:later}
  \end{subfigure}
  \caption{The rewards of trained policy according to the set number of expert trajectories (Average of 5 episodes).}
  \label{fig:reward}
\end{figure*}

\begin{table}[t!]
\footnotesize
        \caption{Performance (Avg. 16 Episodes, 40 Trajectories)}
        \vspace{-3mm}
        \begin{center}
            \scalebox{1}{
            	\begin{tabular}{l|c|c|c}
                    \toprule
                      Average & RAIL (Stacked) & RAIL (Linear) & Expert\\
                    \midrule [1.0pt] 
                    Speed [km/h] & 70.38 & 65.00  &  68.83  \\
                    \# Overtake & 45.04  & 40.03 & 44.48  \\
                    \# Lane change & 15.01 & 13.05 & 14.04  \\
                    Longitudinal  & 2719.38 & 2495.57 & 2642.11\\
                    Lateral  & -122.98 & -175.6 & -132.52\\[0.2ex]
                    \bottomrule
            	\end{tabular}
            	}
        \label{tab:experiment}
        \end{center}
    \vspace{-3mm}
\end{table}

\BfPara{Results}
The purpose of experiments in Fig. \ref{fig:sample} is to show the sample efficiency. 
In order to assess the efficiency, average speed, number of lane changes, number of overtakes, longitudinal reward, and lateral reward were considered as shown in Fig. \ref{fig:sample} and Fig. \ref{fig:reward}.
In Table~\ref{tab:experiment}, it can be seen that the two-layer policy resulted in the highest values of average speed and average overtaking statistics where the values are $70.38$\,km/h and $45.04$, respectively. This is because the trained policies sometimes can achieve higher performance than experts since GAIL-based frameworks perform policy optimization based on the interaction with the environment. On the other hand, the performance of single-layer policy shows $90\%$ performance compared to expert. This is because the single-layer is not enough to handle high dimensional observations properly. Aforementioned, BC tries to minimize 1-step deviation error along the expert demonstration. As a result, the single-layer policy shows undesirable performance due to distribution mismatch between training and testing.
 
In Fig.~\ref{fig:reward}, a longitudinal reward is used to analyze the environmental reward. The longitudinal reward is proportional to the speed; and thus the normalized result shows the same result as the average speed as shown in Fig~\ref{fig:speed}.
In order to assess sensitivity to action decisions, a lateral reward was used. Until the lane change is completed, the host vehicle can change the decision according to the observation. Because the lateral reward occurs continuously during lane change, the frequent changes during the operation lead to reward reduction. 
In Fig.~\ref{fig:later}, the two-layer policy obtains a large lateral reward in the last case. However, the two-layer policy shows more lane changes than the expert. This is because the two-layer policy is less likely to change the decision during the operation. On the other hand, the single-layer policy shows the frequent lane changes than the expert. The single-layer policy obtains the smallest lateral reward. This is because the single-layer policy changes its decision frequently. BC shows the least number of lane changes. However, the trained policy obtains larger reward than the single-layer policy trained by RAIL. The number of lane change is considerably smaller than the single-layer policy; and thus it leads to the trained policy obtains larger lateral reward than the single-layer policy.
The experiment of Fig.~\ref{fig:overtake} was conducted to measure appropriate decisions to imitate the expert demonstration. In order to achieve the similar number of overtakes as the expert, the lane change point and decision should be similar to the expert during the simulation.
In Fig.~\ref{fig:overtake}, the two-layer policy shows the desired performance compared to expert. This result is related to the tendency (i.e., meaningless lane change and decision change) which is shown in Fig.~\ref{fig:lane} and Fig.~\ref{fig:later}. Furthermore, the decision points and actions are similar to the expert. However, the single-layer policy shows a lower number of overtakes than the expert. This is because the average speed is low as well as makes inappropriate lane change decisions based on observation.

In summary, we verified that the proposed RAIL improves the average speed and reduces the number of unnecessary lane changes rather than BC. This means that the RAIL trains driving policies in the correct directions. The experimental results show that the two layer policy achieves desired performance similar to driving experts.


\section{Conclusion}\label{sec:conclusion}
This paper proposes randomized imitation learning (RAIL) for effect autonomous driving policy training which utilizes ADAS functions to guarantee the safety of vehicles. The RAIL is not only a derivative-free but also the simplest model-free reinforcement learning algorithm. Through the proposed algorithm, the policies that successfully drive autonomous vehicles are trained via derivative free optimization. During the training procedure, the simple update step makes the algorithm to be facile; and thus it makes the reconstruction results which get reasonable performance easily. By comparing the performance of the proposed model with complex deep reinforcement learning based methods, we demonstrate that the proposed RAIL trains the policies that achieve desired performance during autonomous driving. This results can be a breakthrough to the common belief that random search in the parameter space of policies can not be competitive in terms of performance. The evaluation results show the possibility that autonomous vehicles can be controlled by the policies which is trained by the proposed RAIL.

\section*{Acknowledgments}
This research was supported by IITP grants funded by the Korea government (MSIT) (No. 2018-0-00170, Virtual Presence in Moving Objects through 5G) and (MSIP) (No. 2017-0-00068, A Development of Driving Decision Engine for Autonomous Driving using Driving Experience Information).
J. Kim is the corresponding author of this paper.
\bibliographystyle{named}
\bibliography{1}

\begin{thebibliography}{}

\bibitem[\protect\citeauthoryear{Ahmed}{1999}]{ahmed1999modeling}
Kazi~I. Ahmed.
\newblock {\em Modeling drivers' acceleration and lane changing behavior}.
\newblock PhD thesis, MIT, 1999.

\bibitem[\protect\citeauthoryear{Finn \bgroup \em et al.\egroup
  }{2016a}]{finn2016connection}
Chelsea Finn, Paul Christiano, Pieter Abbeel, and Sergey Levine.
\newblock A connection between generative adversarial networks, inverse
  reinforcement learning, and energy-based models.
\newblock In {\em NIPS Workshop on Adversarial Training}, 2016.

\bibitem[\protect\citeauthoryear{Finn \bgroup \em et al.\egroup
  }{2016b}]{finn2016guided}
Chelsea Finn, Sergey Levine, and Pieter Abbeel.
\newblock Guided cost learning: {D}eep inverse optimal control via policy
  optimization.
\newblock In {\em ICML}, 2016.

\bibitem[\protect\citeauthoryear{Guo \bgroup \em et al.\egroup }{2018}]{gasil}
Yijie Guo, Junhyuk Oh, Satinder Singh, and Honglak Lee.
\newblock Generative adversarial self-imitation learning.
\newblock {\em arXiv preprint arXiv:1812.00950}, 2018.

\bibitem[\protect\citeauthoryear{Henderson \bgroup \em et al.\egroup
  }{2017a}]{optiongan}
Peter Henderson, Wei-Di Chang, Pierre-Luc Bacon, David Meger, Joelle Pineau,
  and Doina Precup.
\newblock Option{GAN}: Learning joint reward-policy options using generative
  adversarial inverse reinforcement learning.
\newblock {\em arXiv preprint arXiv:1709.06683}, 2017.

\bibitem[\protect\citeauthoryear{Henderson \bgroup \em et al.\egroup
  }{2017b}]{henderson2017deep}
Peter Henderson, Riashat Islam, Philip Bachman, Joelle Pineau, Doina Precup,
  and David Meger.
\newblock Deep reinforcement learning that matters.
\newblock {\em arXiv preprint arXiv:1709.06560}, 2017.

\bibitem[\protect\citeauthoryear{Ho and Ermon}{2016}]{ho2016generative}
Jonathan Ho and Stefano Ermon.
\newblock Generative adversarial imitation learning.
\newblock In {\em NIPS}, 2016.

\bibitem[\protect\citeauthoryear{Hoel \bgroup \em et al.\egroup
  }{2018}]{hoel2018automated}
Carl-Johan Hoel, Krister Wolff, and Leo Laine.
\newblock Automated speed and lane change decision making using deep
  reinforcement learning.
\newblock {\em arXiv preprint arXiv:1803.10056}, 2018.

\bibitem[\protect\citeauthoryear{Islam \bgroup \em et al.\egroup
  }{2017}]{islam2017reproducibility}
Riashat Islam, Peter Henderson, Maziar Gomrokchi, and Doina Precup.
\newblock Reproducibility of benchmarked deep reinforcement learning tasks for
  continuous control.
\newblock {\em arXiv preprint arXiv:1708.04133}, 2017.

\bibitem[\protect\citeauthoryear{Korssen \bgroup \em et al.\egroup
  }{2018}]{korssen2018systematic}
Tim Korssen, Victor Dolk, Joanna van~de Mortel-Fronczak, Michel Reniers, and
  Maurice Heemels.
\newblock Systematic model-based design and implementation of supervisors for
  advanced driver assistance systems.
\newblock {\em IEEE TITS}, 2018.

\bibitem[\protect\citeauthoryear{Kostrikov \bgroup \em et al.\egroup
  }{2019}]{kostrikov2018discriminator}
Ilya Kostrikov, Kumar~Krishna Agrawal, Debidatta Dwibedi, Sergey Levine, and
  Jonathan Tompson.
\newblock Discriminator-actor-critic: Addressing sample inefficiency and reward
  bias in adversarial imitation learning.
\newblock In {\em ICLR}, 2019.

\bibitem[\protect\citeauthoryear{Mania \bgroup \em et al.\egroup }{2018}]{ars}
Horia Mania, Aurelia Guy, and Benjamin Recht.
\newblock Simple random search provides a competitive approach to reinforcement
  learning.
\newblock {\em arXiv preprint arXiv:1803.07055}, 2018.

\bibitem[\protect\citeauthoryear{Mao \bgroup \em et al.\egroup
  }{2017}]{mao2017least}
Xudong Mao, Qing Li, Haoran Xie, Raymond Lau, Zhen Wang, and Stephen~Paul
  Smolley.
\newblock Least squares generative adversarial networks.
\newblock In {\em ICCV}, 2017.

\bibitem[\protect\citeauthoryear{Matyas}{1965}]{matyas1965random}
J~Matyas.
\newblock Random optimization.
\newblock {\em Automation and Remote Control}, 26(2):246--253, 1965.

\bibitem[\protect\citeauthoryear{Min and Kim}{2018}]{min2018deep}
Kyushik Min and Hayoung Kim.
\newblock Deep {Q}-learning based high level driving policy determination.
\newblock In {\em IV}, 2018.

\bibitem[\protect\citeauthoryear{Mnih \bgroup \em et al.\egroup
  }{2015}]{mnih2015human}
Volodymyr Mnih, Koray Kavukcuoglu, David Silver, Andrei~A Rusu, Joel Veness,
  Marc~G Bellemare, Alex Graves, Martin Riedmiller, Andreas~K Fidjeland, Georg
  Ostrovski, et~al.
\newblock Human-level control through deep reinforcement learning.
\newblock {\em Nature}, 2015.

\bibitem[\protect\citeauthoryear{Mukadam \bgroup \em et al.\egroup
  }{2017}]{mukadam2017tactical}
Mustafa Mukadam, Akansel Cosgun, Alireza Nakhaei, and Kikuo Fujimura.
\newblock Tactical decision making for lane changing with deep reinforcement
  learning.
\newblock In {\em NIPS Workshop MLITS}, 2017.

\bibitem[\protect\citeauthoryear{Nagabandi \bgroup \em et al.\egroup
  }{2018}]{nagabandi2018neural}
Anusha Nagabandi, Gregory Kahn, Ronald Fearing, and Sergey Levine.
\newblock Neural network dynamics for model-based deep reinforcement learning
  with model-free fine-tuning.
\newblock In {\em ICRA}, 2018.

\bibitem[\protect\citeauthoryear{Pan \bgroup \em et al.\egroup
  }{2018}]{pan2018agile}
Yunpeng Pan, Ching-An Cheng, Kamil Saigol, Keuntaek Lee, Xinyan Yan, Evangelos
  Theodorou, and Byron Boots.
\newblock Agile autonomous driving using end-to-end deep imitation learning.
\newblock In {\em RSS}, 2018.

\bibitem[\protect\citeauthoryear{Pomerleau}{1989}]{pomerleau1989alvinn}
Dean Pomerleau.
\newblock Alvinn: An autonomous land vehicle in a neural network.
\newblock In {\em NIPS}, 1989.

\bibitem[\protect\citeauthoryear{Pomerleau}{1991}]{pomerleau1991rapidly}
Dean Pomerleau.
\newblock Rapidly adapting artificial neural networks for autonomous
  navigation.
\newblock In {\em NIPS}, 1991.

\bibitem[\protect\citeauthoryear{Rajeswaran \bgroup \em et al.\egroup
  }{2017}]{rajeswaran2017towards}
Aravind Rajeswaran, Kendall Lowrey, Emanuel~V Todorov, and Sham Kakade.
\newblock Towards generalization and simplicity in continuous control.
\newblock In {\em NIPS}, 2017.

\bibitem[\protect\citeauthoryear{Salimans \bgroup \em et al.\egroup
  }{2017}]{es}
Tim Salimans, Jonathan Ho, Xi~Chen, Szymon Sidor, and Ilya Sutskever.
\newblock Evolution strategies as a scalable alternative to reinforcement
  learning.
\newblock {\em arXiv preprint arXiv:1703.03864}, 2017.

\bibitem[\protect\citeauthoryear{Schulman \bgroup \em et al.\egroup
  }{2015}]{schulman2015trust}
John Schulman, Sergey Levine, Pieter Abbeel, Michael Jordan, and Philipp
  Moritz.
\newblock Trust region policy optimization.
\newblock In {\em ICML}, 2015.

\bibitem[\protect\citeauthoryear{Schulman \bgroup \em et al.\egroup
  }{2017}]{schulman2017proximal}
John Schulman, Filip Wolski, Prafulla Dhariwal, Alec Radford, and Oleg Klimov.
\newblock Proximal policy optimization algorithms.
\newblock {\em arXiv preprint arXiv:1707.06347}, 2017.

\bibitem[\protect\citeauthoryear{Silver \bgroup \em et al.\egroup
  }{2016}]{silver2016mastering}
David Silver, Aja Huang, Chris Maddison, Arthur Guez, Laurent Sifre, George Van
  Den~Driessche, Julian Schrittwieser, Ioannis Antonoglou, Veda Panneershelvam,
  Marc Lanctot, et~al.
\newblock Mastering the game of {G}o with deep neural networks and tree search.
\newblock {\em Nature}, 2016.

\bibitem[\protect\citeauthoryear{Ziebart \bgroup \em et al.\egroup
  }{2008}]{ziebart2008maximum}
Brian~D Ziebart, Andrew~L Maas, J~Andrew Bagnell, and Anind~K Dey.
\newblock Maximum entropy inverse reinforcement learning.
\newblock In {\em AAAI}, 2008.

\end{thebibliography}

\end{document}